\documentclass[letterpaper, 10 pt, conference]{ieeeconf}

\IEEEoverridecommandlockouts
\usepackage[T1]{fontenc}
\usepackage{amsmath,amsfonts}
\usepackage{algorithmic}
\usepackage{algorithm}
\usepackage{array}
\usepackage[caption=false,font=normalsize,labelfont=sf,textfont=sf]{subfig}
\usepackage{textcomp}
\usepackage{stfloats}
\usepackage{url}
\usepackage{verbatim}
\usepackage{graphicx}
\usepackage{cite}
\title{\LARGE \bf
Design and Control of a Cable-Driven Switchable Actuator with Torque/Tension Dual Modes for Exoskeletons
}

\author{}

\begin{document}

\title{Design and Control of a Cable-Driven Switchable Actuator with Torque/Tension Dual Modes for Exoskeletons}
\author{Yuanlong Ji, \textit{Student Member, IEEE}, Xu Liu, Xinyuan Cai, Qihan Ye, Xiangyu Xie, Ruizhe Jiang,\\ Shuhan Xiang, Wenjing Liu, Qijun Wang, Yang Chen\textsuperscript{*}, and Xingbang Yang\textsuperscript{*}, \textit{Member, IEEE}
\thanks{\textsuperscript{*}Research supported by the National Natural Science Foundation of China (Grant number 52475291), Beijing Natural Science Foundation (Grant number L222139 and QY26149), and the Fundamental Research Funds for the Central Universities (Grant numbers GW2025-71 and GW2026-MS-39). (Corresponding authors: Yang Chen; Xingbang Yang. E-mails: chenyang4117@163.com; yangxingbang@buaa.edu.cn)}%
\thanks{Yuanlong Ji, Xinyuan Cai, Qihan Ye, Xiangyu Xie, Ruizhe Jiang, Shuhan Xiang, Wenjing Liu, and Xingbang Yang are with the School of Biological Science and Medical Engineering, Beihang University; the Key Laboratory of Biomechanics and Mechanobiology (Beihang University), Ministry of Education; Key Laboratory of Innovation and Transformation of Advanced Medical Devices, Ministry of Industry and Information Technology; National Medical Innovation Platform for Industry-Education Integration in Advanced Medical Devices (Interdiscipline of Medicine and Engineering), Beihang University, Beijing, 100191, China (E-mail: jiyuanlong@buaa.edu.cn).}%
\thanks{Xu Liu and Qijun Wang are with the School of Engineering Medicine, Beihang University, Beijing, 100191, China.}%
\thanks{Yang Chen is with the School of Physics and Mechanical and Electrical Engineering, Longyan University, Longyan, 364012, China.}%
}


\maketitle

\begin{abstract}
Existing wearable exoskeleton architectures are typically constrained by a single mechanical output modality, providing either joint torque around an anatomical joint or linear traction along a limb-training-oriented direction, which limits adaptability to diverse training scenarios. This letter presents a cable-driven switchable actuator (CDSA) that can rapidly switch between torque and tension modes while centralizing all sensing and actuation components at the proximal drive unit. A Coupled Movable Pulley Mechanism (CMPM) provides tension amplification at the distal end-effector, while a bidirectional Cable-Driven Ratchet Mechanism (CDRM) enables mode switching and preload regulation. To eliminate the need for distal instrumentation, multi-source proximal sensors are integrated with a data-driven fusion model to estimate distal output forces. An adaptive dual-mode force control strategy based on iterative learning control (ILC) is further developed. Platform experiments demonstrate transmission efficiencies of 92.4~$\pm$~2.0\% and 96.5~$\pm$~3.3\% in the torque and tension modes, respectively, along with a tension amplification ratio of 2.77~$\pm$~0.10 under tension mode. Tracking tests on simulated knee-joint gait trajectories and short-stroke tension profiles yield stable control, with RMSEs of 4.52~$\pm$~0.51\% and 3.15~$\pm$~0.19\% of the uncontrolled peak value, respectively. Finally, seated human-coupled experiments validate the system's controllable force generation in both joint-torque and linear-traction application modes.
\end{abstract}

\section{INTRODUCTION}

\begin{figure}[t]
    \centering
    \includegraphics[width=\columnwidth]{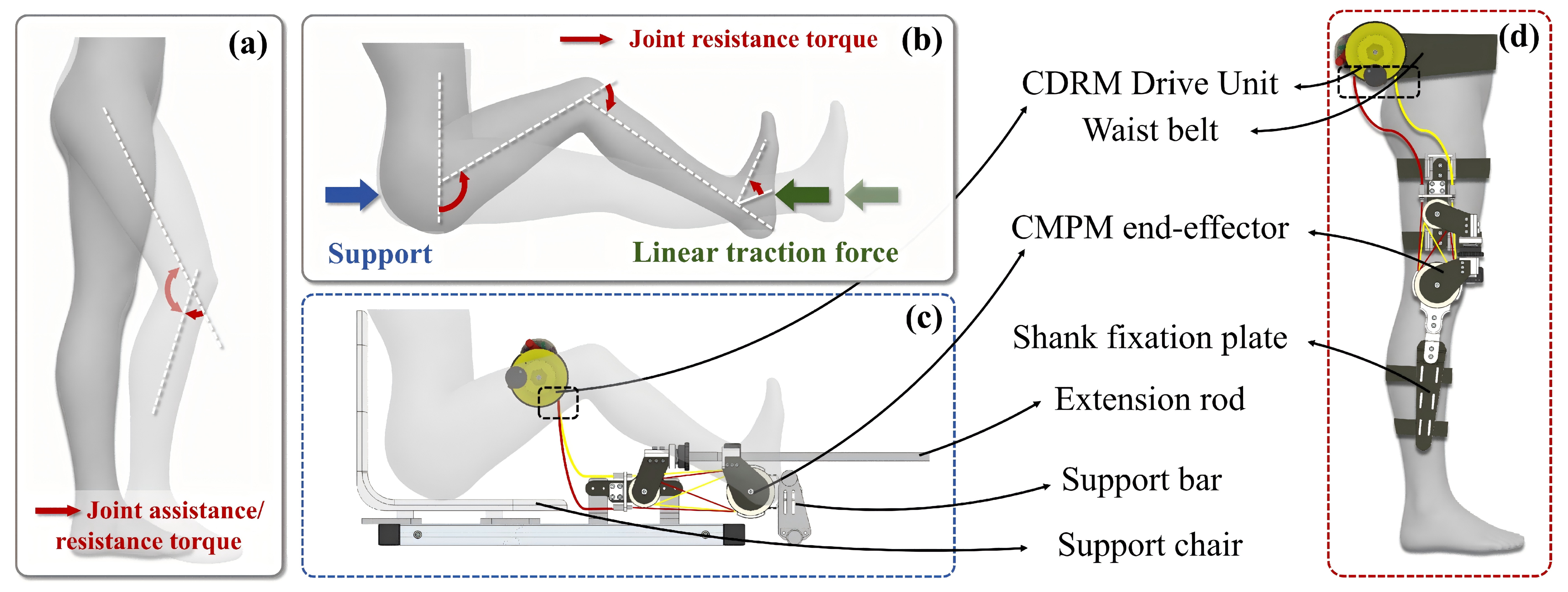}
    \caption{Exoskeleton applications of the CDSA and anchor-point interpretation of the two application-level modes. (a) Joint-torque mode for joint assistance or resistance. (b) Biomechanical analysis of the linear-traction mode. (c) Seated configuration for the linear-traction mode. (d) Wearable configuration for the joint-torque mode. Here, torque and tension modes refer to internal operating states of the CDSA in the actuator-level, whereas joint-torque and linear-traction modes describe how the actuator output is mechanically coupled to the wearer through different anchoring configurations in the application-level.}
    \label{fig_wearable_modes}
\end{figure}

By establishing a direct physical interface with the human body, wearable exoskeletons serve as essential mechanical interventions for both restoring impaired motor functions~\cite{c1,c2,c3} and augmenting physical capacities~\cite{c4,c5,c6}.
In terms of force output strategies, existing exoskeleton
actuation systems are typically limited to a single mechanical
output mode: applying single-DOF torque around a specific
joint axis for direct joint intervention~\cite{c7,c8}, or delivering linear traction along a prescribed direction to the wearer~\cite{c9,c10}. Fig.~\ref{fig_wearable_modes} illustrates representative application instances of the two single-mode mechanical outputs. In the joint-torque configurations (Fig.~\ref{fig_wearable_modes}(a) and Fig.~\ref{fig_wearable_modes}(d)), the reaction load is carried by a local joint frame anchored to the proximal and distal limb segments, thereby transmitting torque about the target joint; in the linear-traction configurations (Fig.~\ref{fig_wearable_modes}(b) and Fig.~\ref{fig_wearable_modes}(c)), the traction force acts between two spatial anchors and induces posture-dependent resistance torques along the limb chain, with the support bar and support chair serving as the two traction anchors in Fig.~\ref{fig_wearable_modes}(c). However, such single-mode interaction falls short of addressing the evolving demands within a single task. In stroke recovery, for instance, the early stage requires localized joint torque to restore precise kinematic patterns~\cite{c11}, whereas the mid-to-late stages demand directional linear traction--akin to elastic band resistance--to facilitate multi-joint loading and comprehensive muscle strengthening~\cite{c12,c13}. Similarly, in sports training, athletes rely on not only joint torque to strengthen targeted muscles or correct specific imbalances~\cite{c14,c15}, but also linear traction to promote multi-joint coordinated exertion and enhance overall functional strength~\cite{c16,c17}. 
Reconfigurable exoskeletons can reduce device cost and space requirements by accommodating different training needs within one platform~\cite{c18}. Similarly, a dual-mode actuator with shared core hardware could eliminate the need for separate task-specific actuation systems, offering a more integrated and cost-effective solution.

The direct-drive joint architecture is widely adopted in wearable robotics for its efficient torque transmission~\cite{c7}. However, due to the rigid rotational coupling between the motor and the joint, it cannot directly generate linear traction, thus requiring additional motion conversion mechanisms or separate drive units. Nevertheless, placing these units on distal limb segments adds mass and inertia, perturbing natural kinematics~\cite{c19,c20,c21}. In contrast, a cable-driven remote actuation architecture circumvents these drawbacks. Decoupling actuation sources from distal joints provides a highly flexible design framework, while reconfigurable pulley mechanisms and diverse cable routing strategies enable efficient conversion of cable tension into either linear traction or joint torque~\cite{c22}. Therefore, cable-driven technology is a competitive solution for constructing the proposed dual-mode exoskeleton actuator.

\begin{figure*}[t]
    \centering
    \includegraphics[width=\textwidth]{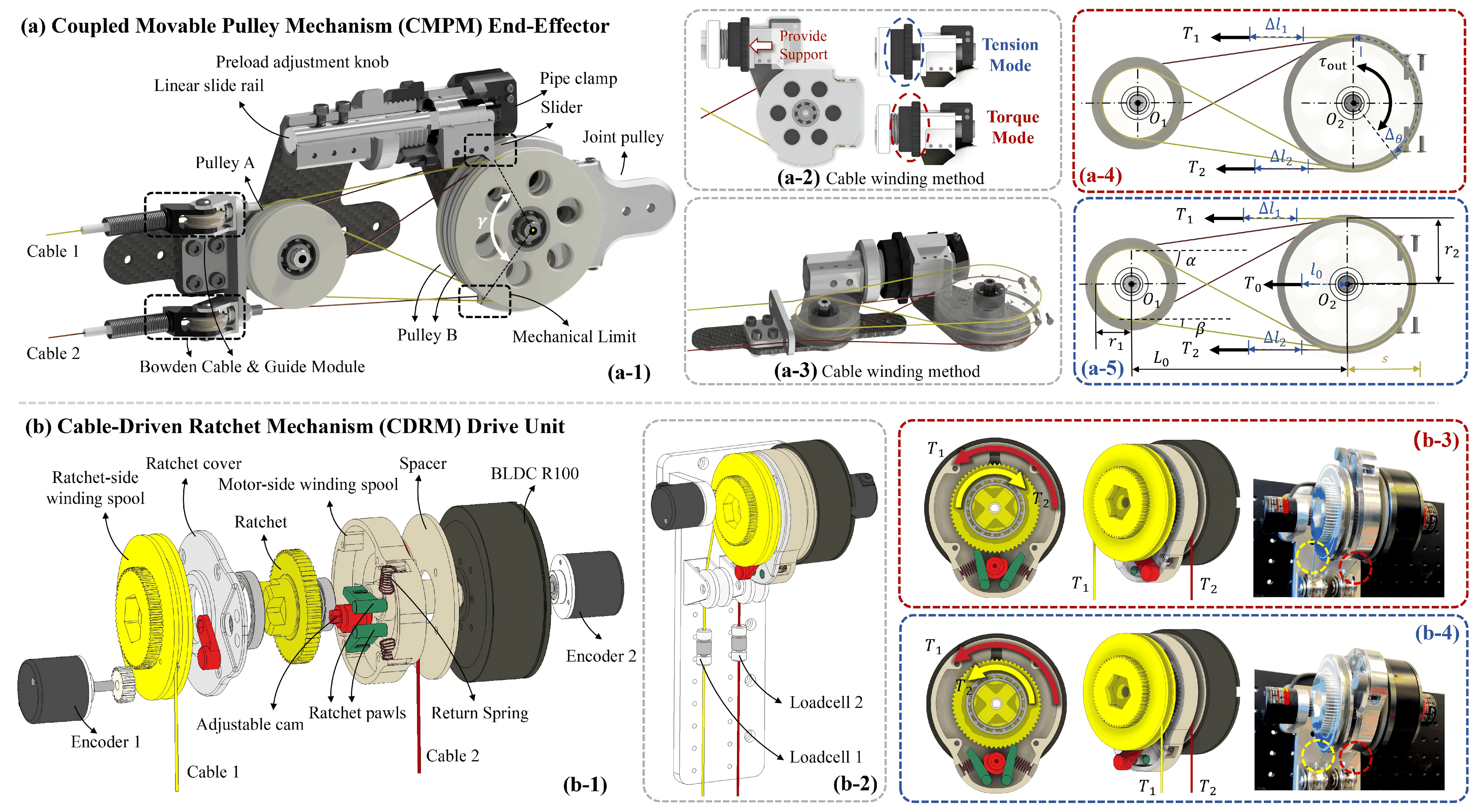}
    \caption{Mechanical structure and working principle of the CDSA. (a) CMPM end-effector: (a-1) structural layout, (a-2) knob states, (a-3) cable winding method, (a-4) torque-mode static analysis, and (a-5) tension-mode static analysis. (b) CDRM drive unit: (b-1) exploded view, (b-2) sensor placement, (b-3) torque-mode winding state, and (b-4) tension-mode winding state.}
    \label{fig_mechanism_details}
\end{figure*}

However, the inherent friction, compliance, and unidirectional force characteristics of cable-driven systems pose significant challenges for dual-mode exoskeleton actuators~\cite{c23,c24}. Structurally, since cables transmit only pulling forces, existing designs typically rely on multi-motor setups~\cite{c22} or electromagnetic clutches~\cite{c27} for mode switching, which compromises system integration. Perceptually, complex nonlinear transmission dynamics hinder the accurate mapping of proximal sensory data to distal outputs~\cite{c28,c29}. While installing force/torque sensors at the distal end improves measurement accuracy, it adds substantial inertia and structural complexity~\cite{c30}. Conversely, analytical models based on Coulomb friction, wrap angle, and equivalent stiffness struggle to adapt to the dynamically time-varying transmission characteristics during operation~\cite{c32}. Hence, data-driven approaches fusing multi-source proximal sensory information have become a promising alternative for distal output estimation~\cite{c33,c34}. In terms of control, inherent compliance and transmission delays tend to induce phase lag and degrade the stability of conventional feedback control~\cite{c40,c41}. Iterative Learning Control (ILC) explicitly exploits the periodic nature of these nonlinearities, leveraging past-cycle error information to preemptively compensate for hysteresis and thereby enhance dynamic tracking performance~\cite{c42}.

To address these challenges--dual-mode actuation, distal force estimation, and tracking control--this paper systematically advances the Coupled Movable Pulley Mechanism (CMPM) originally proposed in~\cite{c22}.
By integrating a mode-switching mechanism into the CMPM and pairing it with a newly designed bidirectional Cable-Driven Ratchet Mechanism (CDRM), the system enables rapid switching between torque and tension modes at the actuator level. To preserve the proximal sensing/actuation layout, we develop an LSTM-based distal-output estimation method using multi-source proximal signals, thus eliminating the need for additional distal force/torque sensors during closed-loop operation. Finally, the control algorithm is optimized by introducing an ILC framework, tailored to cable transmission characteristics, to actively compensate for periodic nonlinearities and time delays inherent in cable transmission.

The remainder of this letter is organized as follows. Section II presents the CDSA design and control framework. Section III describes the platform and human-subject experimental protocols. Section IV reports the experimental results, Section V discusses the findings and limitations, and Section VI concludes this letter.

\section{DESIGN}
This section details the structural design and control algorithms of the CDSA, which synergistically constitute a unified design framework encompassing mechanical implementation, state perception, and closed-loop control.

\subsection{CMPM End-Effector}
The CMPM end-effector weighs only 0.91 kg. As illustrated in Fig.~\ref{fig_mechanism_details}(a), it receives actuation power from the proximal CDRM drive unit via a Bowden cable module. In the tension mode, the preload adjustment knob is rotated to the leftmost position to unlock the degree of freedom (DoF) in the longitudinal direction. This allows the slider and Pulley B to translate freely along a linear guide rail of extendable length, thereby generating a leftward output tension. In the torque mode, by contrast, the preload adjustment knob is repositioned to suppress this linear motion and adjust the cable preload, thereby converting the cable tension into joint torque instead. Simultaneously, driven by the cable, the joint pulley contributes directly to distal torque output. Furthermore, to ensure safety during gait assistance, a mechanical limit slot is integrated into the joint pulley to constrain its travel and prevent excessive displacement.

In the tension mode, the mechanical analysis of the CMPM can be modeled as a combination of two sets of movable pulleys, as illustrated in Fig.~\ref{fig_mechanism_details}(a-5). The displacement lengths of the two cables are denoted by $\Delta l_1$ and $\Delta l_2$, respectively. During the sliding process, the cable inclination angles $\alpha$ and $\beta$ vary dynamically with respect to their linear displacement increment $l_0$. This kinematic relationship can be expressed as:

\begin{equation}
\alpha = \arcsin\left(\frac{r_1 + r_2}{L_0 \pm l_0}\right),\quad
\beta = \arcsin\left(\frac{r_2 - r_1}{L_0 \pm l_0}\right),
\label{eq_alpha}
\end{equation}

\noindent where $L_0$ denotes the initial distance between the centers of rotation of the fixed pulley $O_1$ and the movable pulley $O_2$, as depicted in Fig.~\ref{fig_mechanism_details}(a-5); $r_1$ represents the winding radius of fixed Pulley A; and $r_2$ represents the winding radius of movable Pulley B and the joint pulley. In this operating mode, two cables shorten synchronously, with each movable pulley set comprising three cable segments. Owing to the presence of angles $\alpha$ and $\beta$, as the linear displacement increment $l_0$ approaches an infinitesimal value (i.e., the differential state $\mathrm{d}l_0$), the linear motion reduction ratio $n$ of the CMPM can be expressed as:

\begin{equation}
n = \cos\alpha + \cos\beta + 1.
\label{eq_n}
\end{equation}

\noindent Let $i = 1, 2$ denote the cable indices. The relationship between the linear displacement increment of the CMPM and the shortening lengths of the two cables can be formulated as follows:

\begin{equation}
l_i = n l_0,l_0 \in [0, s],
\label{eq_li}
\end{equation}

\noindent where $s$ is the motion range of Pulley B. The displacement ranges of the two cables can be derived by Equation (4):

\begin{equation}
0 \leq l_i \leq \int_{0}^{s} n\, \mathrm{d}l_0.
\label{eq_li_integral}
\end{equation}

In this mode, the two cables synchronously apply output tensions in the same direction, denoted as $T_1$ and $T_2$. $T_0$ represents the output tension of the CMPM. Based on the principle of virtual work, the static model of the CMPM can be derived as:
\begin{equation}
T_0 = n \cdot (T_1 + T_2).
\label{eq_T0}
\end{equation} 

In the torque mode, as illustrated in Fig.~\ref{fig_mechanism_details}(a-4), the preload adjustment knob locks the position of the slider. Consequently, $\alpha$ and $\beta$ remain constant, while the joint pulley rotates about $O_2$, generating an angular displacement $\Delta\theta$ (defined as positive in the clockwise direction).

\begin{equation}
\Delta\theta = \frac{\Delta l_2 - \Delta l_1}{r_2},
\label{eq_delta_theta}
\end{equation}
\noindent the angular displacement range of the CMPM is denoted as $ \gamma$. In this mode, under the preload exerted by the CDRM, two cables synchronously shorten and lengthen in opposite directions. The displacement ranges of the two cables can be expressed as follows (with the clockwise direction defined as positive):
\begin{equation}
0 \leq l_i \leq \gamma r_2.
\label{eq_li_gamma}
\end{equation}

In the torque mode, two cables execute synchronous movements in opposite directions to regulate the load. The output torque of the CMPM can be expressed as:

\begin{equation}
\tau_0 = r_2 \cdot (T_2 - T_1)
\label{eq_tau0}
\end{equation}
\noindent The static model indicates that the CMPM achieves tension amplification scaled by the linear motion reduction ratio $n$, and torque transmission driven by the differential cable tension. By alternating the dual-cable actuation scheme between co-directional and anti-directional pulling, the transition control between torque and tension outputs can be readily accomplished.

\subsection{CDRM Drive Unit}
By switching between synchronous and differential cable winding, i.e., altering the cable winding direction on the spools, the proximal CDRM drive unit works synergistically with the distal CMPM end-effector to enable torque-tension mode transitions. As depicted in Fig.~\ref{fig_mechanism_details}(b), this drive unit comprises a custom-engineered bidirectional ratchet-based cable routing system and a direct-drive servo motor (Model R100, Nanchang Kude Intelligent Technology Co., Ltd., China).

The bidirectional ratchet-based cable routing system incorporates two winding spools: a motor-side winding spool and a ratchet-side winding spool. These two spools independently anchor Cable 1 and Cable 2, respectively, and are coaxially coupled with two absolute encoders (Shenzhen BriterEncoder Technology Co.,Ltd., China) to measure the extension and retraction of the cables. Furthermore, two load cells (Autoda Intelligent Technology Co., Ltd., China) with a full-scale range of 500 N are integrated into the respective cables to monitor variations in both preload and dynamic tension (Fig.~\ref{fig_mechanism_details}(b-2)). As illustrated in Fig.~\ref{fig_mechanism_details}(b), the ratchet mechanism is mounted between the motor-side winding spool and the ratchet-side winding spool. The ratchet shaft passes through the central bore of the ratchet-side winding spool and mechanically synchronizes with the ratchet-side winding spool. Two symmetrically mirrored pawls, each paired with a return spring, are mounted on the motor-side winding spool. An adjustable cam regulates the self-locking state of the ratchet by modulating the engagement angle of the pawls, thereby switching the force transmission direction of the ratchet-side winding spool, as depicted in Fig.~\ref{fig_mechanism_details}(b). Leveraging the unidirectional self-locking characteristic of the ratchet mechanism, the cable winding direction and the relative rotation angle of the ratchet can be dynamically altered. This enables rapid switching between the torque and tension modes, as well as precise preload adjustment in the torque mode.
Supplementary Video, Part I provides exploded-view animations of the CMPM end-effector and CDRM drive unit, illustrating the structural composition and assembly relationship of the main mechanical components. Supplementary Video, Part II shows the CDRM--CMPM actuator operation, including module connection, preload adjustment, bidirectional torque/tension-mode switching, distal end-effector motion, and motor-driven force-application demonstrations in both operating modes.

To analyze the mode-switching process, a zero-position calibration is first performed after the system is powered on. In this state, Cable 1 is wound out, and the encoder reading is reset to $\theta = 0$, which serves as the reference point for distinguishing between the torque and tension modes. The system then switches between the two modes by changing the ratchet engagement state, winding direction, and winding angle. In the tension mode, the two cables generate co-directional pulling, whereas in the torque mode, they generate anti-directional pulling and an additional winding angle is introduced for preload adjustment. Since the ratchet has a 60-tooth structure, the minimum mechanical step for preload regulation is $6^{\circ}$, indicating that the preload angle can be adjusted in integer multiples of $6^{\circ}$. This design provides a mechanical basis for rapid dual-mode switching and preload regulation in the torque mode.

\begin{figure}[t]
    \centering
    \includegraphics[width=\columnwidth]{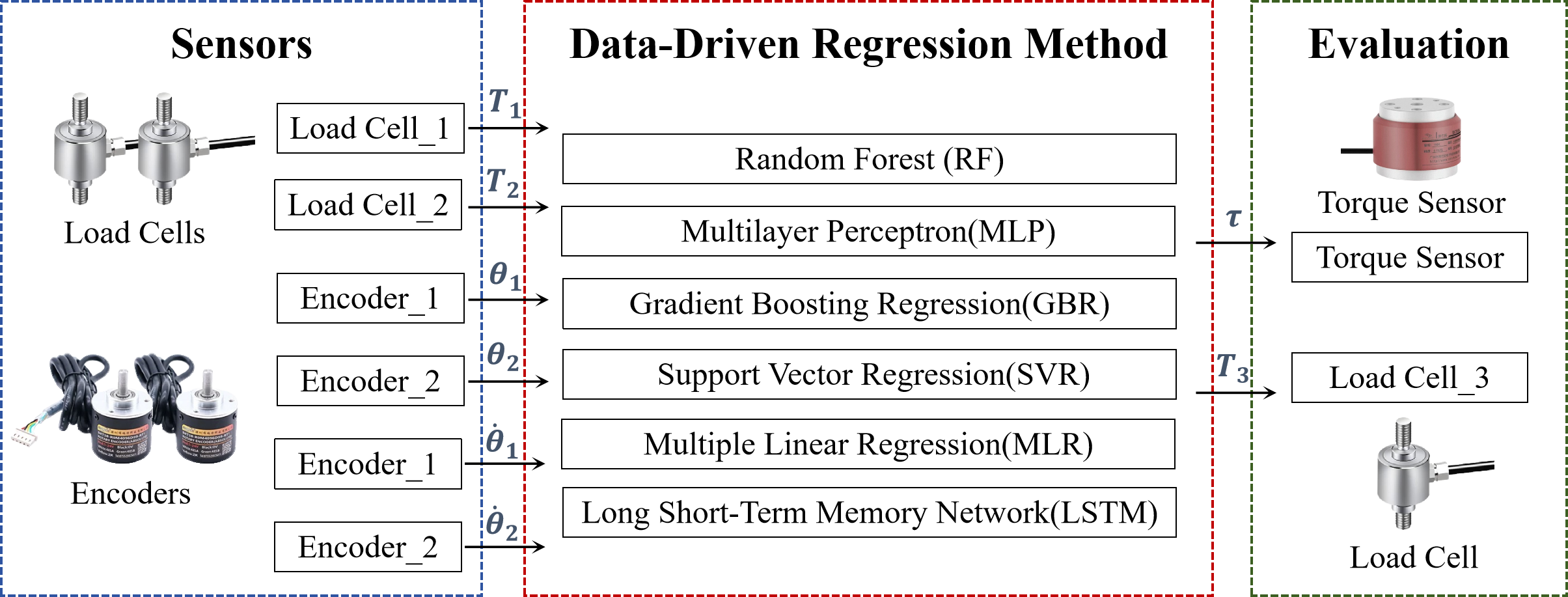}
    \caption{Logic diagram of the multi-sensor fusion algorithm. The blue dashed box encloses the sensors and data forms used for fusion; the red dashed box outlines the data-driven linear regression method; and the green dashed box indicates the sensors reserved for evaluation.}
    \label{fig_experimental_logic}
\end{figure}

\begin{figure}[t]
    \centering
    \includegraphics[width=\columnwidth]{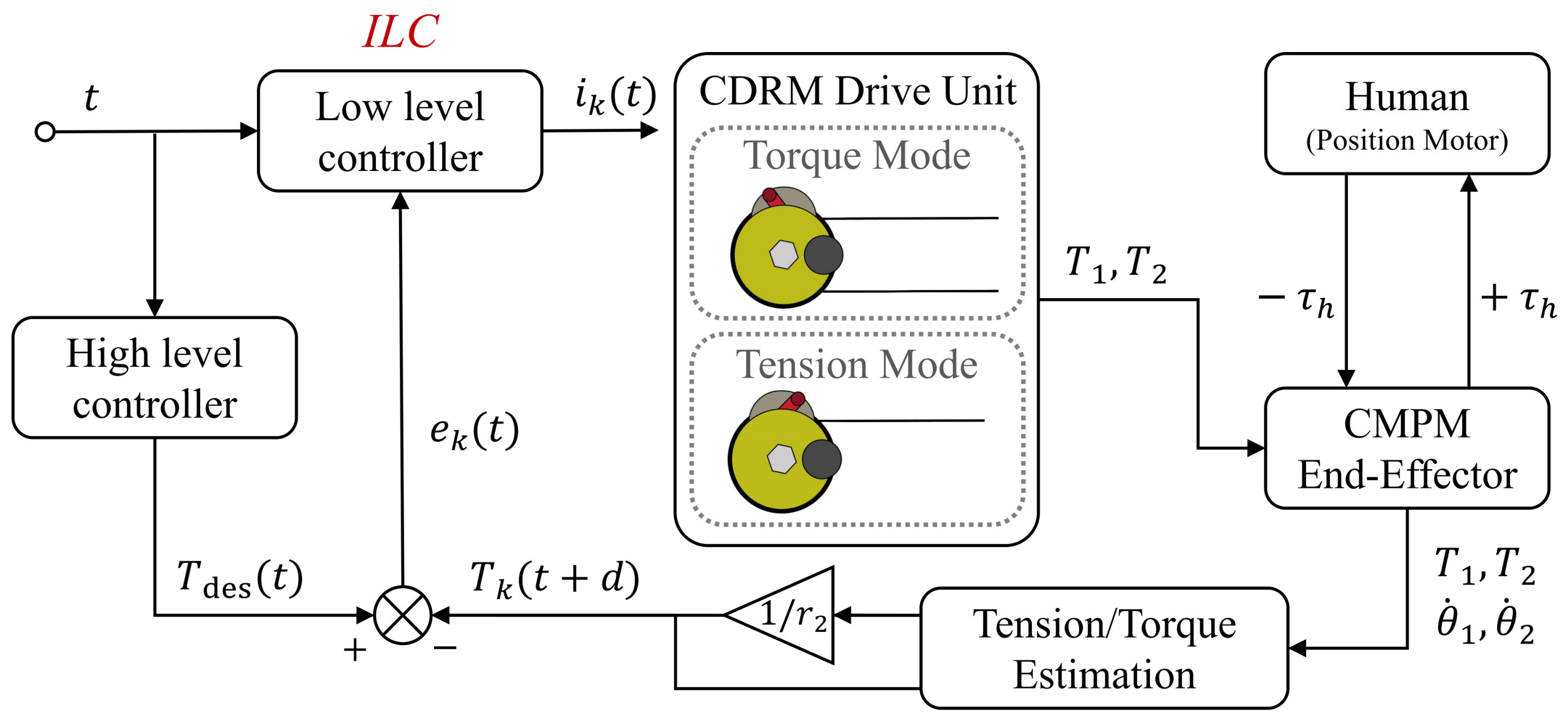}
    \caption{Operational  logic  diagram  of  the  hierarchical  control  framework. The low-level controller employs an iterative learning controller (ILC) for dynamic tracking, while the high-level controller generates the desired distal output(tension or torque)based on task requirements.}
    \label{fig_system_control_block}
\end{figure}

\subsection{Multi-Sensor Fusion Algorithm}
To accurately estimate the distal torque and tension without the necessity of integrating additional sensors at the end-effector, a data-driven regression model is formulated utilizing multi-source sensory information acquired from the CDRM drive unit. As illustrated in Fig.~\ref{fig_experimental_logic}, the overall framework comprises two core modules: multi-source sensory input and distal force regression modeling.

The system acquires three categories of signals: the cable tensions $T_1$ and $T_2$ measured by the load cells, alongside the angular position $\theta_1$, $\theta_2$ and angular velocities $\dot{\theta}_1$, $\dot{\theta}_2$ measured by the encoders. Among these, the tension reflects the actuation effort, whereas the angular position and velocity capture the kinematic and dynamic state of the cables. Based on this rationale, three representative sensing configurations are constructed: tension only, tension combined with angular position, and tension combined with angular velocity. These configurations are designed to assess how the dimensionality of sensory input affects distal force estimation accuracy. During model training, temporarily installed distal torque and tension sensors are used to acquire ground-truth output labels in the torque and tension modes, respectively, thereby constructing the supervised training dataset. Six representative regression methods are selected to predict the distal torque/tension.

\subsection{Iterative Learning Controller}
In each control cycle, the system operates according to the hierarchical control framework shown in Fig.~\ref{fig_system_control_block}. The high-level controller first generates the desired end-effector output based on task requirements. Simultaneously, the proximal tension and encoder measurements are fed into a multi-sensor fusion module to estimate the distal end-effector output in real time for the current cycle. Subsequently, the deviation between the desired command and the estimated output serves as the learning basis for the low-level controller, and its tracking error is defined as:

\begin{equation}
e_k(t)=Q(T_{\mathrm{des}}(t)-T_k(t+d)),
\label{eq_error_detail}
\end{equation}

\noindent where $t$ is the time index, $k$ is the iteration index, $e_k(t)$ is the filtered tracking error, $T_{\mathrm{des}}(t)$ denotes the desired distal output command (torque in torque mode or tension in tension mode), and $T_k(t+d)$ is the estimated distal output after delay alignment. The operator $Q$ is a second-order Butterworth low-pass filter used to suppress high-frequency noise and non-repetitive disturbances. The delay term $d$ compensates for filtering and transmission latency so that the desired command and feedback output are temporally aligned for error calculation. The low-level controller employs an ILC strategy to incrementally correct the control current based on the error signal, and its update law is expressed as:

\begin{equation}
i_{k+1}(t)=i_k(t)+Le_k(t+1),
\label{eq_ilc_update}
\end{equation}

\noindent where $L$ is the learning gain and $i_k(t)$ is the control current applied to the motor during the $k$-th iteration. The updated current command is then fed into the cable-driven actuator, whose motor output is transmitted to the end-effector via the dual-cable mechanism to produce the desired torque or tension. Throughout this process, the human wearer imposes periodic loads, while the multi-sensor fusion model continuously estimates the distal output state and feeds it back to the error calculation stage, thereby closing the control loop.

\subsection{Mode Recognition Algorithm}
To ensure reliable switching between the torque and tension modes, a zero-position calibration is performed after system initialization by winding out Cable 1 and resetting the encoder reading to $\theta=0$. The current mode is then identified directly from the sign of the absolute encoder reading: $\theta>0$ corresponds to the tension mode, whereas $\theta<0$ corresponds to the torque mode. This rule provides a compact and deterministic description of the mode-switching state while avoiding additional distal sensing.

%

The mechanism accurately characterizes the target angle intervals corresponding to different output modes, providing a quantitative description for the mode switching and preload adjustment processes.

\section{EXPERIMENTS}

\subsection{Experimental Setup}

The experimental evaluation comprised two categories: platform experiments and human-subject wearable experiments, both supported by a shared hardware and control system. The platform experiments quantified the intrinsic actuator-level behavior of the CDSA, including static output, distal-output estimation, dynamic tracking, preload sensitivity, and mode switching under controlled boundary conditions. The seated human-subject experiments provided an initial wearable validation of both application-level output modes under human coupling.

\begin{figure*}[t]
    \centering
    \includegraphics[width=\textwidth]{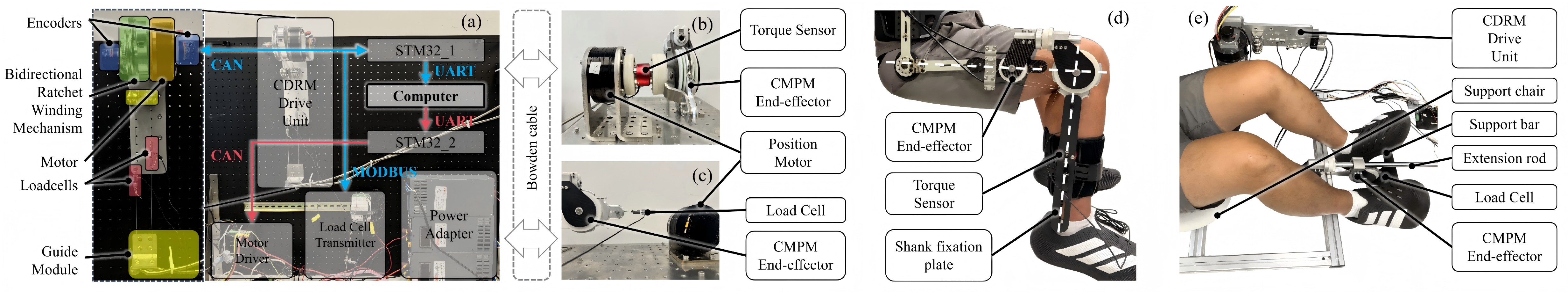}
    \caption{Experimental setup. (a) CDRM hardware and communication layout. (b) Torque-mode setup. (c) Tension-mode setup. (d) Seated human-subject setup for the joint-torque mode. (e) Seated human-subject setup for the linear-traction mode.}
    \label{photo7}
\end{figure*}

For the hardware and control system, Fig.~\ref{photo7}(a) shows the CDRM drive unit, sensors, motor driver, load-cell transmitter, and communication links. STM32\_1 acquired encoder and tension-sensor data via CAN and Modbus, respectively, and transmitted the synchronized data to the host computer through a serial connection. STM32\_2 received control commands from the host computer and relayed them to the motor driver via CAN. The CDRM drive unit was connected to the platform and wearable setups through Bowden cables.

For the platform experiments, two testing platforms were established, as shown in Fig.~\ref{photo7}(b) and Fig.~\ref{photo7}(c), corresponding to the torque mode and tension mode, respectively. Each platform used a position motor (Nanchang Coode Intelligent Technology Co., Ltd.). The torque-mode platform used a torque sensor (Aoda Zhidong Intelligent Technology Co., Ltd.) with a measurement range of 10 N$\cdot$m to measure distal interaction torque, whereas the tension-mode platform used the tension sensor integrated within the actuator to measure distal interaction force.

Three healthy subjects (23.7~$\pm$~2.1 years, 177.3~$\pm$~3.1 cm, 72.7~$\pm$~6.4 kg; mean~$\pm$~SD) completed five trials in each application mode. Approval of all ethical and experimental procedures and protocols was granted by Beihang
University Ethics Committee under No. BM20240132. Written informed consent was obtained from the participant before the experiment. This experiment yielded fifteen ST trials and fifteen SF trials. Subjects maintained the specified posture and actively resisted the applied output. These wearable experiments extend the actuator-level validation to seated human-coupled conditions and demonstrate both joint-torque and linear-traction application modes with the same CDSA hardware.
Both seated tests shared the CDRM drive unit and CMPM end-effector; joint-torque used the knee-mounted configuration, whereas linear traction used support chair/bar anchors.

\subsection{Platform Experiments}

\subsubsection{Static Torque and Tension Output Tests}
To validate torque transmission and tension amplification, the position motor was held fixed while constant current commands of different magnitudes were applied. Proximal and distal sensor data were recorded for 10 torque-mode groups and 6 tension-mode groups.

\subsubsection{Multi-Sensor Fusion Benchmark}
The data-driven fusion approach was evaluated under platform loads representative of human--robot interaction. Motor-current waveform, amplitude, and frequency were varied across startup, shutdown, steady-state, acceleration, deceleration, and oscillatory conditions. The position motor was set to different fixed positions and damping states. The parameter ranges used to collect the training and testing datasets are listed in Table~\ref{table_variables}.

\begin{table}[t]
\caption{Parameter ranges used for multi-sensor fusion dataset collection}
\label{table_variables}
\centering
\resizebox{\columnwidth}{!}{
\begin{tabular}{lcc}
\hline
\textbf{Parameter} & \textbf{Torque Mode} & \textbf{Tension Mode} \\ \hline
\noalign{\vskip 1.5pt}
Current Amplitude & $-30000 \sim 30000$ mA & $0 \sim 10000$ mA \\
Waveform & \multicolumn{2}{c}{sinusoidal, triangular, and square waveforms} \\
Frequency & $0.005 \sim 10$ Hz & $0.005 \sim 2$ Hz \\
Motion Range & $0 \sim 1.05$ rad & $0 \sim 0.08$ m \\ \hline
\end{tabular}
}
\end{table}

\subsubsection{Dynamic Tracking and Preload Tests}
Dynamic platform experiments jointly evaluated tracking under different reference waveforms and cable preloads. The position motor actuated Pulley B to generate prescribed periodic motion: a fitted 10th-order Fourier representation of a normative knee-joint gait profile was used in the torque mode, and a sinusoidal trajectory was used in the tension mode~\cite{c43}. A pretrained long short-term memory (LSTM) regressor combined cable tension and angular velocity to estimate distal output. Sinusoidal, triangular, and square waveforms were used as tracking targets. For the preload comparison, the controller gains, load structure, and desired trajectories were held constant while the ratchet mechanism adjusted cable preload. Each motion profile and each preload level was tested in five repeated trials; Table~\ref{table2} lists the dynamic experimental parameters.

\begin{table}[t]
\caption{Dynamic Experimental Parameters}
\label{table2}
\centering
\resizebox{\columnwidth}{!}{
\begin{tabular}{lcc}
\hline
\textbf{Parameter} & \textbf{Torque Mode} & \textbf{Tension Mode} \\ \hline
Learning Gain $L$ & 100 mA/(N$\cdot$m) & 0.6 mA/N \\
Period & 20.53 s & 4 s \\
Peak Amplitude & 1.12 rad & 0.015 m \\
Discretization Number $N$ & 1000 & 200 \\
Error Alignment Delay $d$ & 20 & 17 \\
Low-pass Cutoff Ratio & 0.2 & 0.2 \\ \hline
\end{tabular}
}
\end{table}

\subsubsection{Mode Recognition and Switching Tests}
The mode recognition performance was evaluated using 10 repeated recognition samples for each operating state. The switching-time experiment consisted of five bidirectional switching cycles between the torque and tension modes. 
Mode switching was manually triggered by the ratchet/cam lever to select the torque- or tension-mode cable path  without the need for complex disassembly of the complete prototype; the single proximal motor then generated output through the selected CDRM-CMPM path.
Because multi-motor and powered clutch-based architectures realize mode availability by adding independent active actuation or switching modules, they are not selected as direct baselines for the repeated cable re-routing task targeted by the proposed single-motor CDRM. Therefore, manual disassembly and reassembly of the same prototype was recorded as a reference for repeated hardware reconfiguration time.

\subsection{Human-Subject Experiments}

The human-subject wearable experiments were conducted to evaluate controllable outputs in both application-level output modes under seated human-coupled conditions. In the seated joint-torque test, each subject wore the actuator according to the configuration shown in Fig.~\ref{photo7}(d). A constant low-amplitude torque reference was applied around the knee joint to emulate controlled static joint loading under wearable human-coupled conditions. The target torque, measured or estimated output torque, and tracking error were recorded to assess output-tracking performance. In the seated linear-traction test, each subject was coupled to the actuator according to the configuration shown in Fig.~\ref{photo7}(e).
A constant target traction force was applied along the prescribed traction direction defined by the designed guide rail to emulate a sustained resistance-like linear-traction load for multi-joint spatial loading and coordinated strengthening. The target force, measured or estimated cable force, and tracking error were recorded to evaluate traction-output tracking under wearable human coupling.

\section{RESULTS}

\subsection{Platform Test Results}

\subsubsection{Static Output Results}
The CDSA produced stable static outputs in both actuator modes.
The torque-mode transmission efficiency reached 92.4~$\pm$~2.0\%, while the tension-mode counterpart achieved 96.5~$\pm$~3.3\%. The measured tension amplification ratio was 2.77~$\pm$~0.10, closely matching the theoretical value of 2.81.

\subsubsection{Multi-Sensor Fusion Results}
\begin{figure}[t]
\centering
\includegraphics[width=\columnwidth]{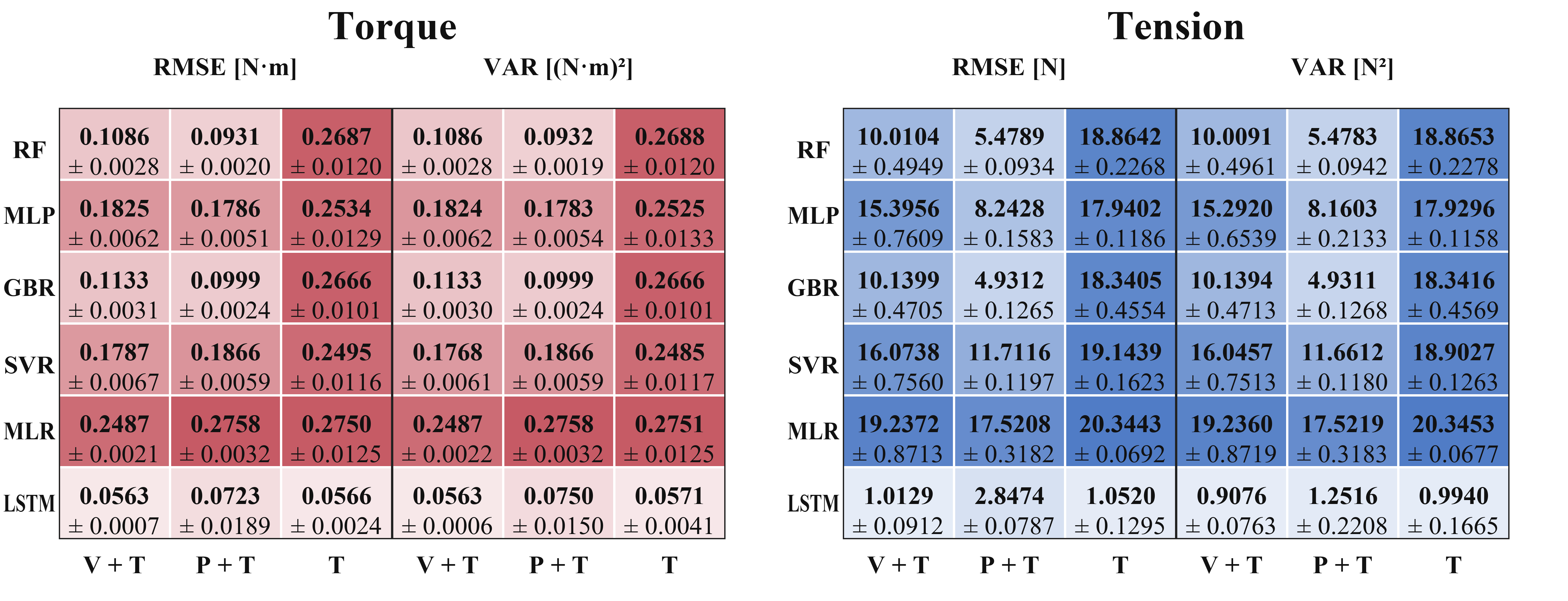}
    \caption{Heatmap-style comparison of torque and tension estimation across sensor configurations and regression algorithms. V+T, P+T, and T denote angular-velocity-plus-tension, angular-position-plus-tension, and tension-only inputs; cells report mean values with variability.}
\label{fig_comparison_results}
\end{figure}

Using the collected datasets, a comparative evaluation was conducted with sensor configuration and regression model as the primary experimental factors. Three sensor input combinations were paired with six regression algorithms to form eighteen experimental conditions, each repeated five times. As shown in Fig.~\ref{fig_comparison_results}, the LSTM model with tension plus angular velocity achieved the lowest numerical errors in both torque and tension modes, with RMSEs of 1.94~$\pm$~0.02\% and 0.57~$\pm$~0.05\%, respectively.
Under the tension-plus-angular-velocity (V+T) input condition, paired $t$-tests showed that LSTM differed significantly from RF in both torque and tension modes ($p<10^{-5}$), whereas the difference between tension-plus-angular-velocity LSTM and tension-only LSTM was not significant in either mode ($p=0.8132$ and $p=0.0848$).

\subsubsection{Dynamic Tracking and Preload Results}
\begin{figure}[t]
\centering
\includegraphics[width=\columnwidth]{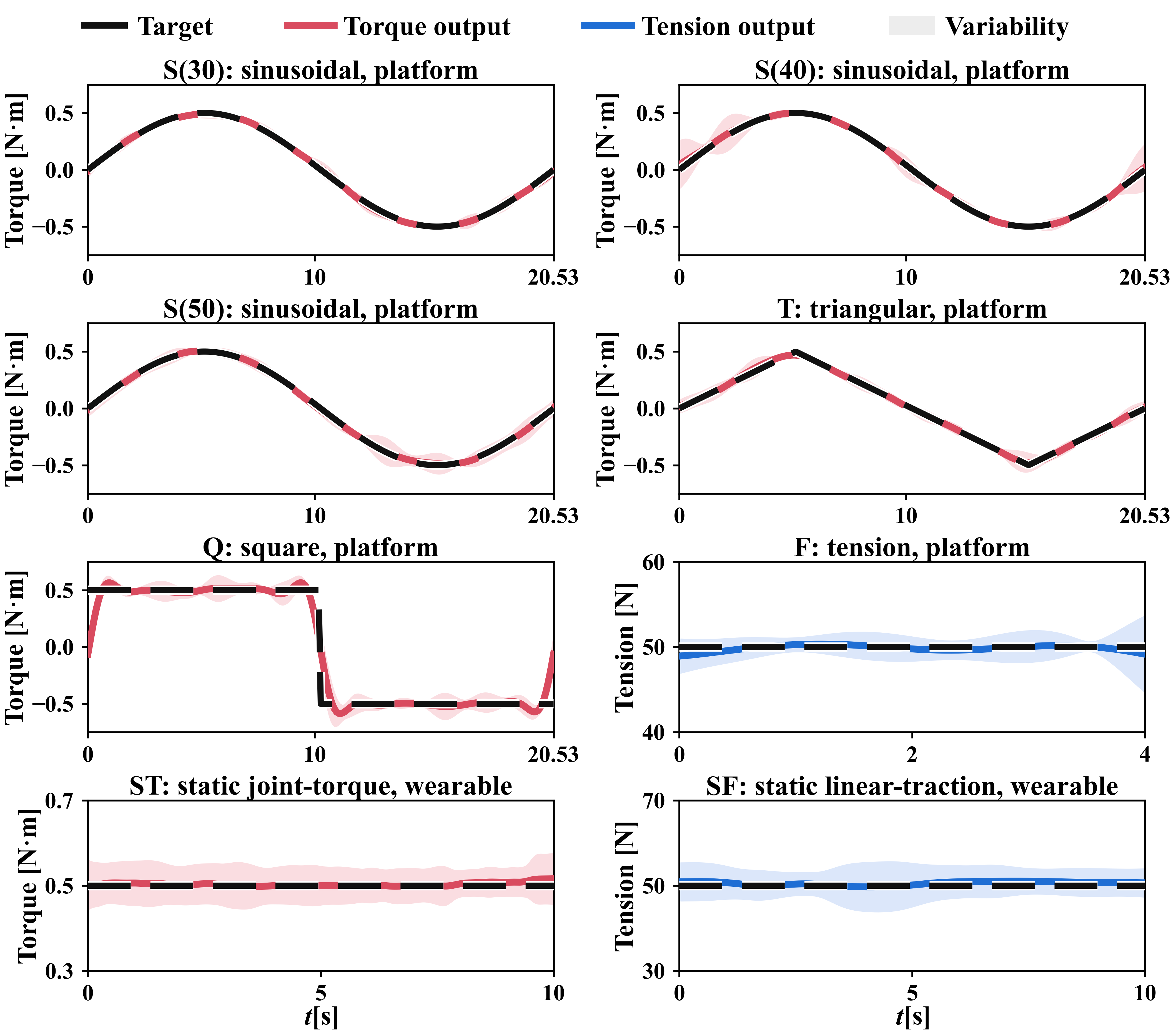}
    \caption{Time-domain tracking performance for platform and seated wearable tests. Dashed lines, solid lines, and shaded regions denote targets, mean outputs, and variability, respectively.}
\label{photo10}
\end{figure}

\begin{figure}[t]
\centering
\includegraphics[width=\columnwidth]{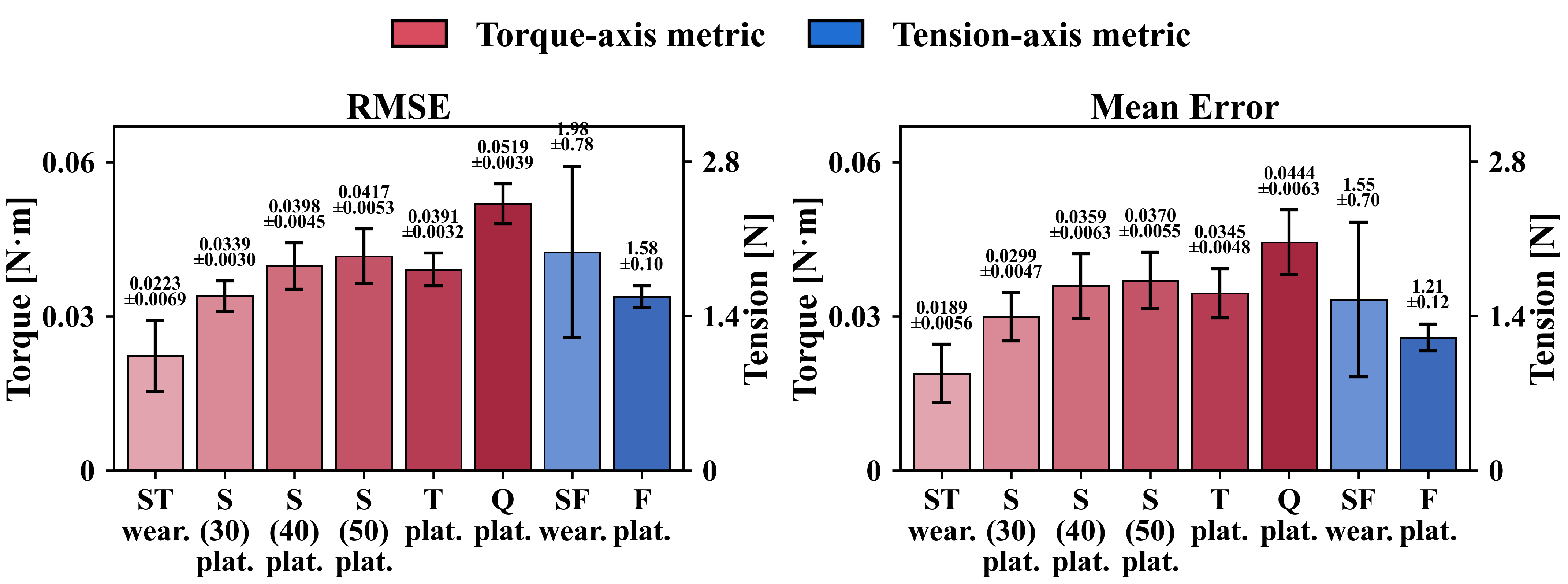}
    \caption{RMSE and mean-error summary. ST/SF: seated static joint-torque/linear-traction tests; S(30)--S(50): sinusoidal torque tracking at 30--50~N preloads; T/Q/F: triangular torque, square torque, and tension tracking. Labels/error bars show means and trial-to-trial SDs.}
\label{photo11}
\end{figure}

The dynamic platform results are shown in Fig.~\ref{photo10} and summarized in Fig.~\ref{photo11}. The ILC controller reproduced the desired output profiles. Torque tracking was most accurate for the triangular wave, with an RMSE of 4.52~$\pm$~0.51\% of the peak unassisted torque and an absolute mean error of 0.036~$\pm$~0.006~N$\cdot$m. Square-wave RMSE remained below 5.71\%, and dynamic tension tracking achieved an RMSE of 3.15~$\pm$~0.19\% relative to the target set point.

A one-way repeated-measures ANOVA showed that normalized RMSE ($p=0.482$) and normalized mean error ($p=0.518$) did not differ significantly across the 30~N, 40~N, and 50~N preload conditions, indicating stable normalized tracking performance over the tested preload range.

\subsubsection{Mode Recognition and Switching Results}
The algorithm correctly identified all repeated samples in both operating states. The ratchet-based mechanism completed mode switching in 13.92~$\pm$~3.96~s; as a reference, manual disassembly and reassembly of the same prototype required 265~$\pm$~37~s.

\subsection{Human-Subject Test Results}

The seated human-subject static-output results are integrated into Fig.~\ref{photo10} and Fig.~\ref{photo11}. The bottom panels of Fig.~\ref{photo10} show the target and measured outputs for the seated static joint-torque and linear-traction tests, demonstrating stable control under human-coupled conditions. 
Fig.~\ref{photo11} reports corresponding ST/SF errors over fifteen trials per mode, with error bars showing trial-to-trial SDs.

\section{DISCUSSION}

The proposed architecture concentrates sensing and actuation in the proximal CDRM drive unit while limiting distal hardware mass to 0.91 kg for the CMPM end-effector, and static tests confirmed stable torque/tension transmission. 
The switching time reported here represents the reconfiguration time of the same prototype via our manually triggered ratchet/cam lever—not a performance comparison with powered clutch- or multi-motor systems. Rather, it quantifies the time saved by our switching strategy relative to complete disassembly/reassembly, while still using a single active motor.

The statistical comparison supports the selection of LSTM as the distal-output estimator. The tension-plus-angular-velocity (V+T) input achieved the lowest numerical errors, while the paired $t$-test showed no significant difference compared with tension-only LSTM. This result suggests that cable tension provides the primary information for distal-output estimation, while encoder-derived angular velocity contributes complementary dynamic-state information related to friction, hysteresis, and cable-motion variations. The fusion of multi-source sensing therefore improves the robustness of output estimation while maintaining a compact sensing architecture. The ILC framework further compensated for periodic transmission errors in both modes, while the preload study indicates that a low but sufficient preload is preferable for balancing cable slack and friction-induced uncertainty.

The seated wearable experiment extends the platform validation to human-coupled conditions, showing that the same CDSA hardware can generate both application-oriented outputs under seated wearable coupling. 
The seated tests share the CDRM drive unit and CMPM end-effector, but linear traction still uses support chair/bar anchors. The current evidence supports actuator-level performance and seated human-coupled output generation. Future work will integrate those anchors into a compact wearable structure inspired by~\cite{c22}, improve CDRM load capacity, and evaluate dynamic gait assistance with larger cohorts. Longer tests will assess attachment stability and switching robustness.

\section{CONCLUSION AND FUTURE WORK}

This paper presents a CDSA that integrates a CMPM end-effector with a bidirectional CDRM drive unit to switch between torque and tension modes while maintaining proximal sensing and actuation. Platform tests validated tension amplification, LSTM-based distal-output estimation from proximal sensing, and ILC-based tracking without additional distal instrumentation, while the seated wearable experiment demonstrated controllable joint-torque and linear-traction outputs under human-coupled conditions. Future work will extend the evaluation to dynamic tasks, larger cohorts, and long-term attachment stability.


\end{document}